\pdfoutput=1

\documentclass[11pt]{article}

\usepackage[final]{acl}

\usepackage{times}
\usepackage{latexsym}

\usepackage[T1]{fontenc}

\usepackage[utf8]{inputenc}

\usepackage{microtype}

\usepackage{inconsolata}

\usepackage{graphicx}
\usepackage{hyperref}
\usepackage{booktabs}
\usepackage{multirow}
\usepackage{stfloats}
\usepackage{amsmath}
\title{FlagEvalMM: A Flexible Framework for Comprehensive Multimodal Model Evaluation}

\author{\textbf{Zheqi He}, \textbf{Yesheng Liu}, \textbf{Jing-shu Zheng}, \textbf{Xuejing Li},\\
         \textbf{Jin-Ge Yao}, \textbf{Bowen Qin}, \textbf{Richeng Xuan}, \textbf{Xi Yang}\\
        BAAI FlagEval Team\\
        \texttt{\{zqhe, yangxi\}@baai.ac.cn}}

\begin{document}
\maketitle
\begin{abstract}
We present \textbf{FlagEvalMM}, an open-source evaluation framework designed to comprehensively assess multimodal models across a diverse range of vision-language understanding and generation tasks, such as visual question answering, text-to-image/video generation, and image-text retrieval. We decouple model inference from evaluation through an independent evaluation service, thus enabling flexible resource allocation and seamless integration of new tasks and models. Moreover, FlagEvalMM utilizes advanced inference acceleration tools (e.g., vLLM, SGLang) and asynchronous data loading to significantly enhance evaluation efficiency. Extensive experiments show that FlagEvalMM offers accurate and efficient insights into model strengths and limitations, making it a valuable tool for advancing multimodal research. The framework is publicly accessible at \href{https://github.com/flageval-baai/FlagEvalMM}{https://github.com/flageval-baai/FlagEvalMM}.
\end{abstract}

\section{Introduction}

With the rapid advancement of large language models (LLMs) \cite{brown2020language}, multimodal models, which integrate multiple forms of input or output data such as text, images, and videos, have experienced significant development in recent years. Currently, vision-language models (VLMs) \cite{openai2023gpt4v, Anthropic2024Claude} are among the most prominent multimodal models. These models typically accept textual and visual inputs—such as images or videos—and generate textual outputs, thus primarily addressing multimodal understanding tasks. Concurrently, text-to-image (T2I) \cite{flux2024,stablediffusion3} and text-to-video (T2V) \cite{kong2024hunyuanvideo,sora} generation tasks, where textual as inputs and generate visual outputs, have also garnered substantial attention, highlighting multimodal generation tasks. Recently, there has been growing interest in developing unified multimodal models capable of integrating both understanding and generation functionalities \cite{chen2025janus,wang2024emu3}.

These developments underscore the need for efficient and comprehensive evaluation frameworks assess multimodal models' diverse capabilities. An ideal evaluation framework should accurately, efficiently, and conveniently assess various capabilities across diverse model architectures. For evaluating VLMs, several frameworks, such as Lmms-Eval \cite{zhang2024lmms} and Vlmevalkit \cite{duan2024vlmevalkit}, have been proposed and widely adopted. Similarly, for evaluating T2I and T2V generation models, CompBench\cite{huang2025t2icompbench++} and VBench \cite{huang2023vbench} are popular choice. However, existing evaluation frameworks typically target specific multimodal tasks, lacking a comprehensive evaluation system capable of supporting a wide array of multimodal tasks uniformly.

Furthermore, current evaluation frameworks generally perform model inference and evaluation within a single runtime environment. With the increasing complexity of evaluation methods, such as use LLM as a judge \cite{gu2024survey}, this architectural choice has revealed several limitations. This tight coupling may lead to conflicts between model inference and evaluation environments, and it can can also impede efficient resource usage.

\begin{figure}[t!]
    \centering
    \includegraphics[width=1.0\linewidth]{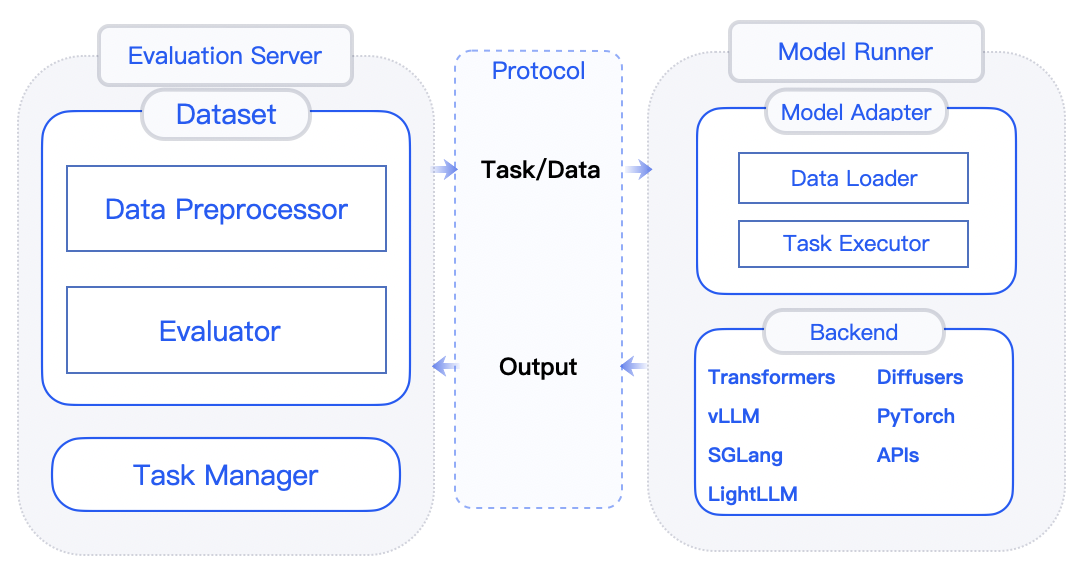}
    \caption{Framework of FlagEvalMM}
    \label{fig:framework}
\end{figure}

In this work, we propose \textbf{FlagEvalMM}, a novel multimodal evaluation framework that addresses existing limitations by decoupling model inference from the evaluation process. As illustrated in Figure \ref{fig:framework}, FlagEvalMM separates the inference environment (\emph{Model Runner}) from an independent evaluation service (\emph{Evaluation Server}). Both components communicate through a lightweight protocol, effectively resolving environment conflicts and enabling more flexible resource allocation. The modular design allows developers to easily add new tasks or models as plugins without modifying the existing framework code. 

Since model inference typically dominates the evaluation time, FlagEvalMM utilizes state-of-the-art inference acceleration libraries (e.g., vLLM \cite{vllm}, SGLang \cite{sglang}, LMDeploy \cite{2023lmdeploy}) to significantly speed up computation. Additionally, it employs asynchronous data loading techniques, such as data prefetching, to further reduce waiting times.

Furthermore, FlagEvalMM provides a comprehensive suite of evaluation paradigms for multimodal understanding and generation tasks, including but not limited to: (a) vision-language understanding (e.g., VQA), (b) text-to-image (T2I) and text-to-video (T2V) generation, and (c) image-text retrieval. Due to its modular architecture, FlagEvalMM easily supports the addition of new task extensions and evaluation metrics, enhancing its versatility and applicability.

To demonstrate its utility, we integrate FlagEvalMM with the Flageval platform\footnote{\url{https://flageval.baai.ac.cn/}} and Huggingface Spaces\footnote{\url{https://huggingface.co/spaces/BAAI/open_flageval_vlm_leaderboard}},enabling users to efficiently deploy new models and conduct comprehensive evaluations. We maintain leaderboards categorized by various multimodal tasks, ranking models according to our meticulously designed capability frameworks. We have cumulatively evaluated hundreds of multimodal models, providing a comprehensive capability analysis of mainstream multimodal models. Our experiments on diverse tasks (vision-language understanding, text-to-image/video generation, and image-text retrieval) highlight the framework's flexibility and extensibility.

In summary, our main contributions are:
\begin{itemize}
    \item We introduce \textbf{FlagEvalMM}, an open-source multimodal evaluation framework that handles both understanding and generation tasks under a unified platform.
    \item By employing a decoupled architecture with an independent evaluation service, FlagEvalMM resolves environment conflicts, enhances flexibility, and improves efficiency in the evaluation process.
    \item We provide extensive empirical results on various tasks, illustrating FlagEvalMM’s capability to deliver detailed insights into different model strengths and limitations.
\end{itemize}
\section{Related Work}
With the significant progress of multimodal models, numerous evaluation frameworks have emerged to assess their capabilities. Specifically, for evaluating vision-language models (VLMs), several benchmarks focus on distinct aspects of performance. For instance, MMMU \cite{yue2023mmmu} evaluates college-level subject knowledge; CMMU \cite{he2024cmmu} assesses Chinese K-12 educational content; Blink \cite{fu2024blink} tests visual perception abilities; MathVerse \cite{zhang2024mathverse} and MathVision \cite{wang2024mathvision} measure mathematical reasoning; OcrBench \cite{liu2024ocrbench} examines text recognition accuracy; and Charxiv \cite{wang2024charxiv} evaluates chart comprehension skills.
 
To facilitate convenient and evaluation across these benchmarks, several evaluation frameworks have been proposed. For instance, Vlmevalkit \cite{duan2024vlmevalkit} is a pioneering open-source multimodal evaluation toolkit. However, its lack of flexibility requires intrusive code modifications for adding new benchmarks or models, making it unsuitable for plug-and-play integrations. VHELM \cite{lee2024vhelm} aggregates multiple datasets to evaluate nine aspects of model performance but suffers from several limitations: first, as an extension of HELM \cite{liang2022holistic}, its architecture is complex, hindering the integration of new models and the expansion of datasets; second, it primarily relies on API calls and has limited support for open-source models. Lmms-Eval \cite{zhang2024lmms}, an excellent and widely-used VLM evaluation framework following the Harness \cite{eval-harness} paradigm, only supports Transformers and vLLM as inference frameworks, thus restricting its flexibility. Furthermore, it does not accommodate evaluations of multimodal generation tasks, limiting its applicability to unified multimodal models.

Regarding the evaluation of multimodal generation tasks, benchmarks are fewer, and the evaluation methods, especially for image or video outputs, are inherently more complex. HEIM \cite{heim} is a comprehensive framework for evaluating text-to-image generation, but similar to VHELM, it is built upon HELM and presents usability challenges. VBench \cite{huang2023vbench} systematically evaluates video generative models across 16 hierarchical and disentangled dimensions, yet it is exclusively tailored to video generation tasks. In contrast to these existing frameworks, our proposed FlagEvalMM offers enhanced flexibility and ease of use, supporting a wide range of multimodal understanding and generation tasks through a unified, user-friendly interface.

\section{System Design}
\label{sec:protocol}
In this section, we present the system design of our proposed framework, FlagEvalMM. As illustrated in Figure \ref{fig:framework}, the system comprises two main components: an evaluation server and a model runner, which communicate through a carefully designed protocol via HTTP. The demonstration video of FlagEvalMM is available is available online.\footnote{Video available at: \url{https://youtu.be/L7EtacjoM0k}}
We will discuss the design of each component in detail.
\subsection{Evaluation Server}

\begin{figure*}[t!]
    \centering
    \includegraphics[width=0.9\linewidth]{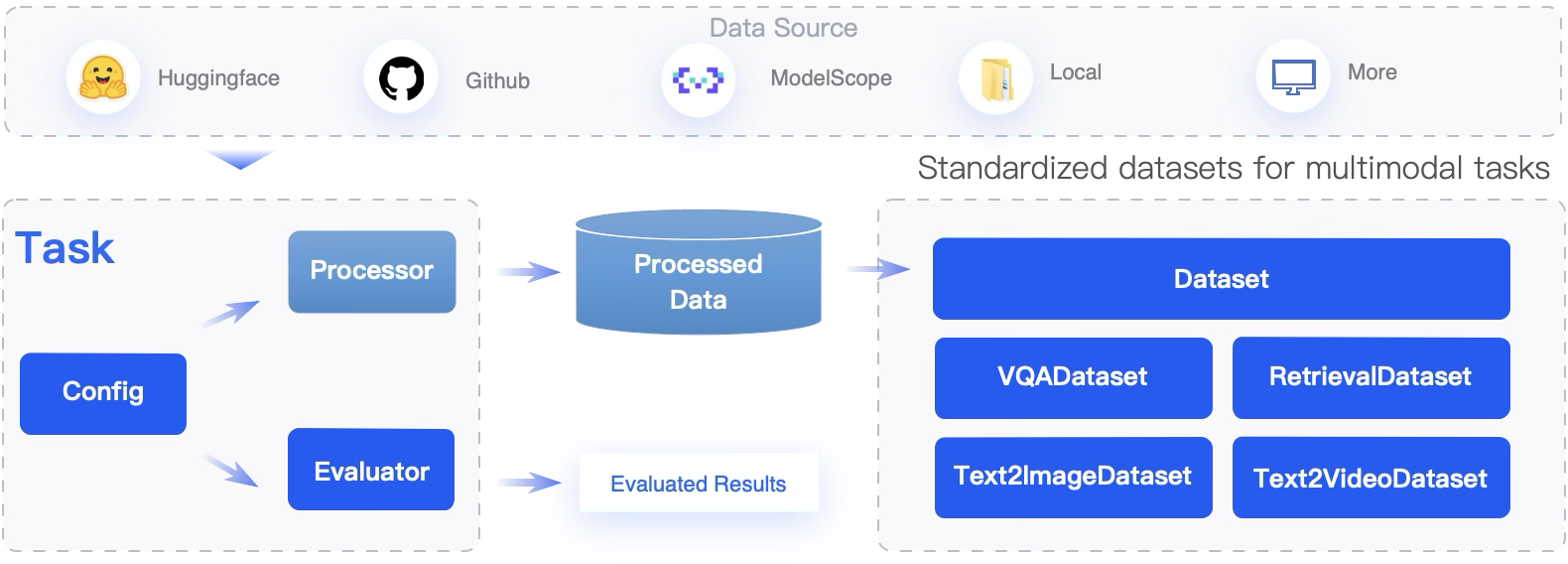}
    \caption{Components and workflow of the evaluation server}
    \label{fig:eval_server}
\end{figure*}

As illustrated in Figure \ref{fig:eval_server}, the evaluation server provides data to the model runner and evaluates model performance. A \textbf{Task} serves as the smallest executable unit within the evaluation server, consisting of three core components:

\begin{itemize} 
\item \textbf{Processor}: Performs data preprocessing, converting datasets from various sources into a standardized format, stored persistently. 
\item \textbf{Config}: Provides configuration parameters such as evaluation metrics and prompt template. 
\item \textbf{Evaluator}: Evaluates model outputs and generates performance metrics. \end{itemize}

The workflow for each task is as follows: read configurations to acquire metadata, distribute data to models, await model outputs, and finally evaluate the generated results. The evaluation server is designed with scalability in mind and can be deployed on cloud platforms to decouple evaluation and inference. While predefined Dataset types and Evaluators are provided, users can also define and register customized Datasets and Evaluators for specific tasks.

\subsection{Model Runner}
The Model Runner is responsible for executing model inference, offering significant flexibility while following the defined Communication Protocol with the evaluation server (see Section \ref{sec:protocol}). As illustrated in the right part of Figure \ref{fig:framework}, the Model Runner consists of two primary components: the \textbf{Model Adapter} and the \textbf{Backend}.

Model adapter plays as the bridge between the evaluation server and the model inference backend, It fetches data from the evaluation server, schedules tasks, and invokes backend processes for model inference. For convenience, commonly used model adapters are provided within our model zoo, including support for OpenAI-style REST API, and popular services such as Gemini and Anthropic (further details are provided in Appendix \S \ref{sec:commercial_api}). Users may directly utilize these predefined adapters or develop custom adapters tailored to their specific requirements. 

The \textbf{Backend} is the inference engine responsible for executing the model computation, user can choose the backend according to their own needs.To optimize inference efficiency, FlagEvalMM officially supports high-performance backends like vLLM, SGLang and MLDeploy. Alternatively, users can directly leverage popular libraries, such as Transformers, Diffusers, PyTorch, or other APIs for inference. To enhance efficiency and reduce redundant computations, we implement a caching mechanism based on SQLite \cite{gaffney2022sqlite}, a lightweight database system. When caching is enabled, the system computes a hash value for input data (including text, images, and parameters) and uses this hash as a unique key to store inference results. Subsequent identical requests retrieve the stored results directly from the cache, significantly reducing processing overhead.

\subsection{Communication Protocol}

\begin{figure*}[t]
    \centering
    \includegraphics[width=1\linewidth]{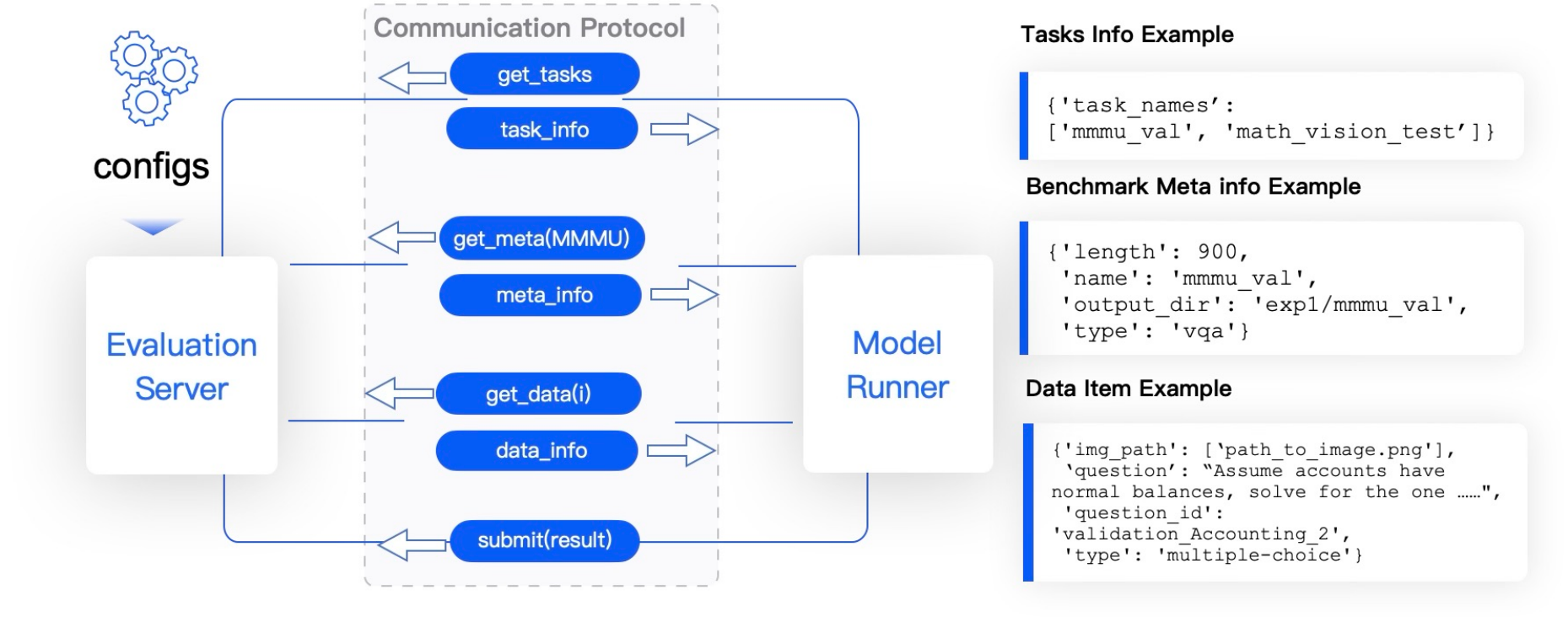}
    \caption{Communication protocol between evaluation server and model runner}
    \label{fig:protocol}
\end{figure*}

The communication protocol between the evaluation server and the model runner is designed to be simple, modular, and extensible. As illustrated in Figure \ref{fig:protocol}, the protocol supports the complete evaluation lifecycle, including task retrieval, metadata provisioning, data access, and result submission. All interactions between the evaluation server and model runner adhere to a RESTful HTTP pattern \cite{fielding2000architectural}, with each evaluation step corresponding to a dedicated API.

The protocol starts with the model runner requesting the available tasks via \texttt{get\_tasks}, and then querying detailed task information with \texttt{task\_info}. After selecting a task, the runner retrieves task-level metadata \texttt{meta\_info} using \texttt{get\_meta}. These metadata include the number of samples, task type (e.g., VQA, T2I), output directory, and other necessary settings.

Once the task setup is complete, the model runner requests specific evaluation items using the \texttt{get\_data(i)}.The returned \texttt{data\_info} includes necessary details like image paths, textual prompts, and unique question identifiers. After inference, the runner submits its predictions back to the evaluation server via the \texttt{submit(result)}.

Each step in the communication protocol supports distributed and parallelized model evaluation. The protocol's modular design also enables easy integration of new task types or data formats without requiring modifications to the core communication logic. As a result, FlagEvalMM remains flexible and easily adaptable to various multimodal evaluation scenarios.

\begin{table*}[htbp]
\centering
\begin{tabular}{l|ccc|ccccc}
\toprule
\multirow{3}{*}{Model} & \multicolumn{3}{c|}{Average Rank} & \multicolumn{5}{c}{Capability Score} \\
\cmidrule(lr){2-4} \cmidrule(lr){5-9}
 & Overall & EN & ZH & Gen & Math & Chart & Vis & Text \\
\midrule
gemini-2.0-pro & 2.1 & 2.4 & 1.5 & 64.00 & 52.18 & 67.06 & 62.73 & 78.22 \\
Qwen2.5-VL-72B & 4.6 & 5.4 & 2.5 & 61.30 & 35.45 & 67.00 & 60.90 & 77.63 \\
Qwen2.5-VL-32B & 6.7 & 7.8 & 4.0 & 60.17 & 42.57 & 62.15 & 59.22 & 74.68 \\
claude-3-7-sonnet-20250219 & 6.9 & 4.2 & 13.5 & 58.98 & 49.31 & 71.19 & 66.55 & 67.69 \\
InternVL2\_5-78B & 6.9 & 7.2 & 6.0 & 61.31 & 37.80 & 60.14 & 62.97 & 70.87 \\
gpt-4o-2024-11-20 & 8.1 & 7.2 & 10.5 & 58.39 & 30.82 & 65.50 & 62.02 & 70.31 \\
claude-3-5-sonnet-20241022 & 8.1 & 6.2 & 13.0 & 59.14 & 45.24 & 71.89 & 62.66 & 67.00 \\
Qwen2-VL-72B & 10.4 & 12.2 & 6.0 & 57.30 & 32.53 & 60.06 & 54.48 & 71.75 \\
gemini-1.5-pro & 11.0 & 8.0 & 18.5 & 53.29 & 50.80 & 62.41 & 56.74 & 63.62 \\
Mistral-Small-3.1-24B & 12.6 & 9.6 & 20.0 & 53.36 & 32.40 & 64.94 & 60.49 & 62.25 \\
llava-onevision-qwen2-72b & 20.3 & 18.0 & 26.0 & 45.84 & 32.90 & 52.09 & 48.55 & 49.48 \\
Molmo-72B-0924 & 22.0 & 18.8 & 30.0 & 43.27 & 26.31 & 54.27 & 50.12 & 44.98 \\
\bottomrule
\end{tabular}
\caption{Ability evaluation of some frontier VLM models. For Gen (General Knowledge), Math, Chart, Vis (Visual Perception), Text (Text Recognition and Understanding), scores are averaged across English and Chinese evaluations.}
\label{tab:vlm_abilities}
\end{table*}

\section{Evaluation Results and Analysis}
We have evaluate more than 50 multimodal understanding models and 30 multimodal generation models on the FlagevalMM leaderboard. In this paper, we focus on the performance of VLMs and text-to-image models. we select some frontier models from various companies and research institutions for detailed analysis.

\subsection{Datasets and Evaluation Metrics}
\subsubsection{Multimodal Understanding}
To comprehensively evaluate the multimodal understanding capabilities of models and address the dataset contamination and metric saturation issues \cite{mmstar}, we selected multiple recent public and self-constructed evaluation datasets for this VLM assessment: Charxiv \cite{wang2024charxiv}, CII-Bench \cite{zhang2024mllmsunderstanddeepimplication}, CMMMU \cite{zhang2024cmmmu}, MMMU \cite{yue2023mmmu}, MMMU-Pro \cite{yue2024mmmu}, MathVision \cite{wang2024mathvision}, MathVerse \cite{zhang2024mathverse}, MMVET-v2 \cite{yu2024mm}, Blink \cite{fu2024blink}, and self-constructed subjective image-text QA dataset and text recognition and understanding dataset. These datasets cover five capabilities: general knowledge, mathematical, chart comprehension, visual perception, and text recognition and understanding, dach dataset can be mapped to one or more capabilities Additionally, we distinguished between Chinese and English capabilities based on question language and cultural type. 

Except for the two self-constructed benchmarks, all datasets are publicly available academic datasets. Public datasets utilized the default prompts and accuracy calculation methods provided by their original sources. The self-constructed subjective evaluation dataset employs binary manual scoring to judge correctness. The self-constructed text recognition and understanding evaluation adopts the automatic accuracy evaluation method from OCRBench \cite{liu2024ocrbench}, determining correctness based on whether the manually annotated standard answer string is a subsequence of the model's response.

\subsubsection{Multimodal Generation}
\label{sec:multimodal_generation}
For multimodal generation tasks, we evaluate the result for 4 aspects: consistency with the prompt, realism, aesthetic quality, and safety. In FlagevalMM, we currently support several metrics for automatic evaluation of multimodal generation models. In our leaderboard, we combined some automatic evaluation metrics with human evaluation to provide a more comprehensive evaluation, we choose VQAScore \cite{lin2024evaluating}, Q-Align \cite{wu2024q}, VideoScore \cite{he2024videoscore} as the automatic evaluation metrics. In human evaluation, we employs 3 human evaluators to score the image in 4 aspects above, and the final score is the average of the 3 human scores. The detailed annotation guideline can be found in Appendix \S \ref{app:human_evaluation}.

Beyond standard datasets like COCO \cite{lin2014microsoft} and GenAI Bench \cite{li2024genai} available in FlagEvalMM, our leaderboard uses self-constructed datasets for text-to-image and text-to-video tasks. The text-to-image dataset contains 414 self-designed high-quality prompts, while the text-to-video dataset includes 148 prompts (100 self-designed, 48 public). The self-constructed datasets are evaluated using the same automatic evaluation metrics as the public datasets.

\subsection{Leaderboard}
In this section, we present evaluation results for representative state-of-the-art multimodal models.

\subsubsection{Results of VLMs}

Table \ref{tab:vlm_abilities} summarizes the performance of representative VLMs across five key multimodal capabilities. The left side of the table shows the overall average rank along with language-specific average ranks, while the right side details capability scores, each representing averages from evaluations conducted in both English and Chinese. Models are ranked based on their overall average rank.

Our analysis reveals substantial progress among recent open-source VLMs. Notably, the Qwen2.5 series \cite{Qwen2025Qwen2.5VL} surpasses several earlier commercial models, highlighting significant advancements within the open-source community. 
This improvement indicates a narrowing performance gap between open-source and proprietary solutions in multimodal understanding tasks. 
However, some models, such as Mistral-3.1\cite{mistralsmall31} and Claude 3.7 \cite{Anthropic2025Claude}, exhibit pronounced performance discrepancies across different languages and cultural contexts, performing notably better in English than in Chinese. These results underscore persistent challenges regarding cross-lingual and cross-cultural generalization in current VLM architectures. According to some case study, we found VLMs currently exhibit instability and inaccuracies in tasks involving spatial reasoning, position estimation, and counting. Additionally, they struggle with classic computer vision challenges such as occlusion, varying illumination, deformation, and perspective changes.

\subsubsection{Results of text-to-image models}
\begin{table*}[ht]
\centering
\resizebox{\textwidth}{!}{
\begin{tabular}{l|c|cccc|ccc}
\toprule
\multirow{2}{*}{Model} & \multirow{2}{*}{Weighted} & \multicolumn{4}{c|}{Human Evaluation} & \multicolumn{3}{c}{Automated Evaluation} \\
\cline{3-9}
& & Cons & Real & Aes & Safety & VQAS & OA-Qua & OA-Aes \\
\midrule
Hunyuan-Image & 73.00 & 67.93 & 66.67 & 78.50 & 100.00 & 73.76 & 95.36 & 81.00 \\
Doubao-Image v2.1 & 71.74 & 69.79 & 61.90 & 75.00 & 94.64 & 76.69 & 89.96 & 73.24 \\
DALL-E 3 & 70.12 & 70.24 & 57.51 & 68.38 & 98.21 & 81.82 & 94.42 & 89.92 \\
Kolors & 68.80 & 68.53 & 62.43 & 63.84 & 92.86 & 75.60 & 88.60 & 80.77 \\
FLUX.1 schnell & 68.39 & 61.95 & 64.34 & 73.18 & 99.11 & 77.95 & 93.53 & 74.60 \\
Firefly Image 3 & 66.15 & 62.80 & 57.07 & 68.90 & 95.54 & 74.39 & 88.92 & 76.91 \\
Midjourney v6.1 & 65.91 & 67.56 & 46.95 & 64.58 & 98.21 & 77.63 & 86.82 & 77.60 \\
Stable Diffusion 3.5 Large & 65.22 & 67.86 & 45.61 & 60.86 & 100.00 & 78.28 & 89.47 & 73.47 \\
CogView-3 Plus & 64.34 & 67.63 & 45.68 & 57.37 & 99.11 & 80.16 & 90.72 & 80.15 \\
\bottomrule
\end{tabular}
}
\caption{Performance comparison of text-to-image models across human and automated evaluation metrics.}
\label{tab:t2i_comparison}
\end{table*}

Table \ref{tab:t2i_comparison} compares the performance of selected text-to-image models using both human and automated evaluation metrics. Since some T2I models only support English prompts, the results presented in the table are based on a subset of English prompts. Models are ranked according to the weighted average of human evaluation scores.

The results demonstrate that commercial models, such as Hunyuan-Image \cite{hunyanimage} and Doubao-Image \cite{doubaoimage}, generally achieve higher performance in human evaluation compared to open-source counterparts like FLUX \cite{flux2024} and CogView-3 \cite{zheng2024cogview3}. Notably, while automated metrics offer useful insights, they do not always align closely with human judgments. For instance, in the consistency dimension, the VQAScore exhibits a Pearson correlation coefficient \cite{cohen2009pearson} of only \textit{0.76} with human evaluation scores. Similarly, for aesthetic quality, the OneAlign-Aesthetic metric yields a moderate correlation of \textit{0.59}. These observations highlight the limitations of current automated evaluation methods and suggest the necessity for further refinement to better reflect human perception. According to some case study, we found that T2I models often struggle with generating high-quality images for human motion scenarios and accurately depicting specified object.
\section{Conclusion and Future Work}
In this work, we introduce \textbf{FlagEvalMM}, an open-source integrates both multimodal understanding and generation tasks within a unified platform. By decoupling model inference from the evaluation process, FlagEvalMM effectively mitigates environmental conflicts and significantly enhances flexibility. Moreover, integration with public platforms such as FlagEval and Huggingface Spaces further enhances ease of use and accessibility. In the future, we plan to incorporate additional evaluation methodologies, such as multi-round conversational tasks, interactive gameplay with vision-language models, and advanced reasoning capability assessments. These extensions aim to broaden the scope and depth of FlagEvalMM.

\section*{Limitations}
Due to the rapidly evolution of evaluation methods and models, our work integrates only a selected subset of existing evaluation approaches and benchmarks. Additionally, a significant gap remains between automated evaluation and human assessment in generation tasks, necessitating continued reliance on manual evaluation. 
\section*{Acknowledgments}
This work was supported by the National Science and Technology Major Project of China (Grant No. 2022ZD0116306). We thank Xue Sun for his valuable contribution in  adding numerous new datasets to our open-source project.
\bibliography{flagevalmm}

\begin{thebibliography}{47}
\providecommand{\natexlab}[1]{#1}

\bibitem[{AI(2025)}]{mistralsmall31}
Mistral AI. 2025.
\newblock \href {https://mistral.ai/news/mistral-small-3-1} {Mistral small 3.1}.

\bibitem[{Anthropic(2024)}]{Anthropic2024Claude}
Anthropic. 2024.
\newblock Claude 3.5 sonnet.
\newblock \url{https://www.anthropic.com/news/claude-3-5-sonnet}.
\newblock Accessed: 2025-01-18.

\bibitem[{Anthropic(2025)}]{Anthropic2025Claude}
Anthropic. 2025.
\newblock Claude 3.7 sonnet.
\newblock \url{https://www.anthropic.com/news/claude-3-7-sonnet}.
\newblock Accessed: 2025-03-08.

\bibitem[{Brown et~al.(2020)Brown, Mann, Ryder, Subbiah, Kaplan, Dhariwal, Neelakantan, Shyam, Sastry, Askell et~al.}]{brown2020language}
Tom Brown, Benjamin Mann, Nick Ryder, Melanie Subbiah, Jared~D Kaplan, Prafulla Dhariwal, Arvind Neelakantan, Pranav Shyam, Girish Sastry, Amanda Askell, and 1 others. 2020.
\newblock Language models are few-shot learners.
\newblock \emph{Advances in neural information processing systems}, 33:1877--1901.

\bibitem[{ByteDance(2024)}]{doubaoimage}
ByteDance. 2024.
\newblock Doubao image.
\newblock \url{https://www.volcengine.com/docs/6791/1366783}.

\bibitem[{Chen et~al.(2024)Chen, Li, Dong, Zhang, Zang, Chen, Duan, Wang, Qiao, Lin et~al.}]{mmstar}
Lin Chen, Jinsong Li, Xiaoyi Dong, Pan Zhang, Yuhang Zang, Zehui Chen, Haodong Duan, Jiaqi Wang, Yu~Qiao, Dahua Lin, and 1 others. 2024.
\newblock Are we on the right way for evaluating large vision-language models?
\newblock In \emph{The Thirty-eighth Annual Conference on Neural Information Processing Systems}.

\bibitem[{Chen et~al.(2025)Chen, Wu, Liu, Pan, Liu, Xie, Yu, and Ruan}]{chen2025janus}
Xiaokang Chen, Zhiyu Wu, Xingchao Liu, Zizheng Pan, Wen Liu, Zhenda Xie, Xingkai Yu, and Chong Ruan. 2025.
\newblock Janus-pro: Unified multimodal understanding and generation with data and model scaling.
\newblock \emph{arXiv preprint arXiv:2501.17811}.

\bibitem[{Cohen et~al.(2009)Cohen, Huang, Chen, Benesty, Benesty, Chen, Huang, and Cohen}]{cohen2009pearson}
Israel Cohen, Yiteng Huang, Jingdong Chen, Jacob Benesty, Jacob Benesty, Jingdong Chen, Yiteng Huang, and Israel Cohen. 2009.
\newblock Pearson correlation coefficient.
\newblock \emph{Noise reduction in speech processing}, pages 1--4.

\bibitem[{Contributors(2023)}]{2023lmdeploy}
LMDeploy Contributors. 2023.
\newblock Lmdeploy: A toolkit for compressing, deploying, and serving llm.
\newblock \url{https://github.com/InternLM/lmdeploy}.

\bibitem[{Contributors(2024)}]{sglang}
SGLang Contributors. 2024.
\newblock \href {https://github.com/sgl-project/sglang} {Sglang: A fast serving framework for large language models and vision language models}.
\newblock Accessed: 2025-03-23.

\bibitem[{Duan et~al.(2024)Duan, Yang, Qiao, Fang, Chen, Liu, Dong, Zang, Zhang, Wang et~al.}]{duan2024vlmevalkit}
Haodong Duan, Junming Yang, Yuxuan Qiao, Xinyu Fang, Lin Chen, Yuan Liu, Xiaoyi Dong, Yuhang Zang, Pan Zhang, Jiaqi Wang, and 1 others. 2024.
\newblock Vlmevalkit: An open-source toolkit for evaluating large multi-modality models.
\newblock In \emph{Proceedings of the 32nd ACM international conference on multimedia}, pages 11198--11201.

\bibitem[{Esser et~al.(2024)Esser, Kulal, Blattmann, Entezari, M{\"u}ller, Saini, Levi, Lorenz, Sauer, Boesel et~al.}]{stablediffusion3}
Patrick Esser, Sumith Kulal, Andreas Blattmann, Rahim Entezari, Jonas M{\"u}ller, Harry Saini, Yam Levi, Dominik Lorenz, Axel Sauer, Frederic Boesel, and 1 others. 2024.
\newblock Scaling rectified flow transformers for high-resolution image synthesis.
\newblock In \emph{Forty-first international conference on machine learning}.

\bibitem[{Fielding(2000)}]{fielding2000architectural}
Roy~Thomas Fielding. 2000.
\newblock \emph{Architectural styles and the design of network-based software architectures}.
\newblock University of California, Irvine.

\bibitem[{Fu et~al.(2024)Fu, Hu, Li, Feng, Wang, Lin, Roth, Smith, Ma, and Krishna}]{fu2024blink}
Xingyu Fu, Yushi Hu, Bangzheng Li, Yu~Feng, Haoyu Wang, Xudong Lin, Dan Roth, Noah~A Smith, Wei-Chiu Ma, and Ranjay Krishna. 2024.
\newblock Blink: Multimodal large language models can see but not perceive.
\newblock In \emph{European Conference on Computer Vision}, pages 148--166. Springer.

\bibitem[{Gaffney et~al.(2022)Gaffney, Prammer, Brasfield, Hipp, Kennedy, and Patel}]{gaffney2022sqlite}
Kevin~P Gaffney, Martin Prammer, Larry Brasfield, D~Richard Hipp, Dan Kennedy, and Jignesh~M Patel. 2022.
\newblock Sqlite: past, present, and future.
\newblock \emph{Proceedings of the VLDB Endowment}, 15(12).

\bibitem[{Gao et~al.(2024)Gao, Tow, Abbasi, Biderman, Black, DiPofi, Foster, Golding, Hsu, Le~Noac'h, Li, McDonell, Muennighoff, Ociepa, Phang, Reynolds, Schoelkopf, Skowron, Sutawika, Tang, Thite, Wang, Wang, and Zou}]{eval-harness}
Leo Gao, Jonathan Tow, Baber Abbasi, Stella Biderman, Sid Black, Anthony DiPofi, Charles Foster, Laurence Golding, Jeffrey Hsu, Alain Le~Noac'h, Haonan Li, Kyle McDonell, Niklas Muennighoff, Chris Ociepa, Jason Phang, Laria Reynolds, Hailey Schoelkopf, Aviya Skowron, Lintang Sutawika, and 5 others. 2024.
\newblock \href {https://doi.org/10.5281/zenodo.12608602} {A framework for few-shot language model evaluation}.

\bibitem[{Gu et~al.(2024)Gu, Jiang, Shi, Tan, Zhai, Xu, Li, Shen, Ma, Liu et~al.}]{gu2024survey}
Jiawei Gu, Xuhui Jiang, Zhichao Shi, Hexiang Tan, Xuehao Zhai, Chengjin Xu, Wei Li, Yinghan Shen, Shengjie Ma, Honghao Liu, and 1 others. 2024.
\newblock A survey on llm-as-a-judge.
\newblock \emph{arXiv preprint arXiv:2411.15594}.

\bibitem[{He et~al.(2024{\natexlab{a}})He, Jiang, Zhang, Ku, Soni, Siu, Chen, Chandra, Jiang, Arulraj et~al.}]{he2024videoscore}
Xuan He, Dongfu Jiang, Ge~Zhang, Max Ku, Achint Soni, Sherman Siu, Haonan Chen, Abhranil Chandra, Ziyan Jiang, Aaran Arulraj, and 1 others. 2024{\natexlab{a}}.
\newblock Videoscore: Building automatic metrics to simulate fine-grained human feedback for video generation.
\newblock In \emph{Proceedings of the 2024 Conference on Empirical Methods in Natural Language Processing}, pages 2105--2123.

\bibitem[{He et~al.(2024{\natexlab{b}})He, Wu, Zhou, Xuan, Liu, Yang, Zhu, and Huang}]{he2024cmmu}
Zheqi He, Xinya Wu, Pengfei Zhou, Richeng Xuan, Guang Liu, Xi~Yang, Qiannan Zhu, and Hua Huang. 2024{\natexlab{b}}.
\newblock Cmmu: a benchmark for chinese multi-modal multi-type question understanding and reasoning.
\newblock In \emph{Proceedings of the Thirty-Third International Joint Conference on Artificial Intelligence}, pages 830--838.

\bibitem[{Huang et~al.(5555)Huang, Duan, Sun, Xie, Li, and Liu}]{huang2025t2icompbench++}
Kaiyi Huang, Chengqi Duan, Kaiyue Sun, Enze Xie, Zhenguo Li, and Xihui Liu. 5555.
\newblock \href {https://doi.ieeecomputersociety.org/10.1109/TPAMI.2025.3531907} {{T2I-CompBench++: An Enhanced and Comprehensive Benchmark for Compositional Text-to-Image Generation }}.
\newblock \emph{IEEE Transactions on Pattern Analysis Machine Intelligence}, pages 1--17.

\bibitem[{Huang et~al.(2024)Huang, He, Yu, Zhang, Si, Jiang, Zhang, Wu, Jin, Chanpaisit, Wang, Chen, Wang, Lin, Qiao, and Liu}]{huang2023vbench}
Ziqi Huang, Yinan He, Jiashuo Yu, Fan Zhang, Chenyang Si, Yuming Jiang, Yuanhan Zhang, Tianxing Wu, Qingyang Jin, Nattapol Chanpaisit, Yaohui Wang, Xinyuan Chen, Limin Wang, Dahua Lin, Yu~Qiao, and Ziwei Liu. 2024.
\newblock {VBench}: Comprehensive benchmark suite for video generative models.
\newblock In \emph{Proceedings of the IEEE/CVF Conference on Computer Vision and Pattern Recognition}.

\bibitem[{Kong et~al.(2024)Kong, Tian, Zhang, Min, Dai, Zhou, Xiong, Li, Wu, Zhang et~al.}]{kong2024hunyuanvideo}
Weijie Kong, Qi~Tian, Zijian Zhang, Rox Min, Zuozhuo Dai, Jin Zhou, Jiangfeng Xiong, Xin Li, Bo~Wu, Jianwei Zhang, and 1 others. 2024.
\newblock Hunyuanvideo: A systematic framework for large video generative models.
\newblock \emph{arXiv preprint arXiv:2412.03603}.

\bibitem[{Kwon et~al.(2023)Kwon, Li, Zhuang, Sheng, Zheng, Yu, Gonzalez, Zhang, and Stoica}]{vllm}
Woosuk Kwon, Zhuohan Li, Siyuan Zhuang, Ying Sheng, Lianmin Zheng, Cody~Hao Yu, Joseph~E. Gonzalez, Hao Zhang, and Ion Stoica. 2023.
\newblock Efficient memory management for large language model serving with pagedattention.
\newblock In \emph{Proceedings of the ACM SIGOPS 29th Symposium on Operating Systems Principles}.

\bibitem[{Labs(2024)}]{flux2024}
Black~Forest Labs. 2024.
\newblock Flux.
\newblock \url{https://github.com/black-forest-labs/flux}.

\bibitem[{Lee et~al.(2024)Lee, Tu, Wong, Zheng, Zhou, Mai, Roberts, Yasunaga, Yao, Xie et~al.}]{lee2024vhelm}
Tony Lee, Haoqin Tu, Chi~Heem Wong, Wenhao Zheng, Yiyang Zhou, Yifan Mai, Josselin Roberts, Michihiro Yasunaga, Huaxiu Yao, Cihang Xie, and 1 others. 2024.
\newblock Vhelm: A holistic evaluation of vision language models.
\newblock \emph{Advances in Neural Information Processing Systems}, 37:140632--140666.

\bibitem[{Lee et~al.(2023)Lee, Yasunaga, Meng, Mai, Park, Gupta, Zhang, Narayanan, Teufel, Bellagente et~al.}]{heim}
Tony Lee, Michihiro Yasunaga, Chenlin Meng, Yifan Mai, Joon~Sung Park, Agrim Gupta, Yunzhi Zhang, Deepak Narayanan, Hannah Teufel, Marco Bellagente, and 1 others. 2023.
\newblock Holistic evaluation of text-to-image models.
\newblock \emph{Advances in Neural Information Processing Systems}, 36:69981--70011.

\bibitem[{Li et~al.(2024)Li, Lin, Pathak, Li, Fei, Wu, Ling, Xia, Zhang, Neubig et~al.}]{li2024genai}
Baiqi Li, Zhiqiu Lin, Deepak Pathak, Jiayao Li, Yixin Fei, Kewen Wu, Tiffany Ling, Xide Xia, Pengchuan Zhang, Graham Neubig, and 1 others. 2024.
\newblock Genai-bench: Evaluating and improving compositional text-to-visual generation.
\newblock \emph{arXiv preprint arXiv:2406.13743}.

\bibitem[{Liang et~al.(2022)Liang, Bommasani, Lee, Tsipras, Soylu, Yasunaga, Zhang, Narayanan, Wu, Kumar et~al.}]{liang2022holistic}
Percy Liang, Rishi Bommasani, Tony Lee, Dimitris Tsipras, Dilara Soylu, Michihiro Yasunaga, Yian Zhang, Deepak Narayanan, Yuhuai Wu, Ananya Kumar, and 1 others. 2022.
\newblock Holistic evaluation of language models.
\newblock \emph{arXiv preprint arXiv:2211.09110}.

\bibitem[{Lin et~al.(2014)Lin, Maire, Belongie, Hays, Perona, Ramanan, Doll{\'a}r, and Zitnick}]{lin2014microsoft}
Tsung-Yi Lin, Michael Maire, Serge Belongie, James Hays, Pietro Perona, Deva Ramanan, Piotr Doll{\'a}r, and C~Lawrence Zitnick. 2014.
\newblock Microsoft coco: Common objects in context.
\newblock In \emph{Computer vision--ECCV 2014: 13th European conference, zurich, Switzerland, September 6-12, 2014, proceedings, part v 13}, pages 740--755. Springer.

\bibitem[{Lin et~al.(2024)Lin, Pathak, Li, Li, Xia, Neubig, Zhang, and Ramanan}]{lin2024evaluating}
Zhiqiu Lin, Deepak Pathak, Baiqi Li, Jiayao Li, Xide Xia, Graham Neubig, Pengchuan Zhang, and Deva Ramanan. 2024.
\newblock Evaluating text-to-visual generation with image-to-text generation.
\newblock In \emph{European Conference on Computer Vision}, pages 366--384. Springer.

\bibitem[{Liu et~al.(2024)Liu, Li, Huang, Yang, Yu, Li, Yin, Liu, Jin, and Bai}]{liu2024ocrbench}
Yuliang Liu, Zhang Li, Mingxin Huang, Biao Yang, Wenwen Yu, Chunyuan Li, Xu-Cheng Yin, Cheng-Lin Liu, Lianwen Jin, and Xiang Bai. 2024.
\newblock Ocrbench: on the hidden mystery of ocr in large multimodal models.
\newblock \emph{Science China Information Sciences}, 67(12):220102.

\bibitem[{OpenAI(2023)}]{openai2023gpt4v}
OpenAI. 2023.
\newblock \href {https://cdn.openai.com/papers/GPTV_System_Card.pdf} {Gpt-4v(ision) system card}.
\newblock \emph{OpenAI Research}.

\bibitem[{OpenAI(2024)}]{sora}
OpenAI. 2024.
\newblock \href {https://openai.com/index/sora-system-card/} {Sora}.

\bibitem[{Team.(2025)}]{Qwen2025Qwen2.5VL}
Qwen Team. 2025.
\newblock \href {https://arxiv.org/abs/2502.13923} {Qwen2.5-vl technical report}.
\newblock \emph{Preprint}, arXiv:2502.13923.

\bibitem[{Tencent(2024)}]{hunyanimage}
Tencent. 2024.
\newblock Hunyuan image.
\newblock \url{https://cloud.tencent.com/document/product/1729/105969}.

\bibitem[{Wang et~al.(2024{\natexlab{a}})Wang, Pan, Shi, Lu, Ren, Zhou, Zhan, and Li}]{wang2024mathvision}
Ke~Wang, Junting Pan, Weikang Shi, Zimu Lu, Houxing Ren, Aojun Zhou, Mingjie Zhan, and Hongsheng Li. 2024{\natexlab{a}}.
\newblock \href {https://openreview.net/forum?id=QWTCcxMpPA} {Measuring multimodal mathematical reasoning with math-vision dataset}.
\newblock In \emph{The Thirty-eight Conference on Neural Information Processing Systems Datasets and Benchmarks Track}.

\bibitem[{Wang et~al.(2024{\natexlab{b}})Wang, Zhang, Luo, Sun, Cui, Wang, Zhang, Wang, Li, Yu et~al.}]{wang2024emu3}
Xinlong Wang, Xiaosong Zhang, Zhengxiong Luo, Quan Sun, Yufeng Cui, Jinsheng Wang, Fan Zhang, Yueze Wang, Zhen Li, Qiying Yu, and 1 others. 2024{\natexlab{b}}.
\newblock Emu3: Next-token prediction is all you need.
\newblock \emph{arXiv preprint arXiv:2409.18869}.

\bibitem[{Wang et~al.(2024{\natexlab{c}})Wang, Xia, He, Chen, Liu, Zhu, Liang, Wu, Liu, Malladi et~al.}]{wang2024charxiv}
Zirui Wang, Mengzhou Xia, Luxi He, Howard Chen, Yitao Liu, Richard Zhu, Kaiqu Liang, Xindi Wu, Haotian Liu, Sadhika Malladi, and 1 others. 2024{\natexlab{c}}.
\newblock Charxiv: Charting gaps in realistic chart understanding in multimodal llms.
\newblock \emph{Advances in Neural Information Processing Systems}, 37:113569--113697.

\bibitem[{Wu et~al.(2024)Wu, Zhang, Zhang, Chen, Liao, Li, Gao, Wang, Zhang, Sun et~al.}]{wu2024q}
Haoning Wu, Zicheng Zhang, Weixia Zhang, Chaofeng Chen, Liang Liao, Chunyi Li, Yixuan Gao, Annan Wang, Erli Zhang, Wenxiu Sun, and 1 others. 2024.
\newblock Q-align: teaching lmms for visual scoring via discrete text-defined levels.
\newblock In \emph{Proceedings of the 41st International Conference on Machine Learning}, pages 54015--54029.

\bibitem[{Yu et~al.(2024)Yu, Yang, Ren, Li, Wang, Lin, Lin, Liu, Wang, and Wang}]{yu2024mm}
Weihao Yu, Zhengyuan Yang, Linfeng Ren, Linjie Li, Jianfeng Wang, Kevin Lin, Chung-Ching Lin, Zicheng Liu, Lijuan Wang, and Xinchao Wang. 2024.
\newblock Mm-vet v2: A challenging benchmark to evaluate large multimodal models for integrated capabilities.
\newblock \emph{arXiv preprint arXiv:2408.00765}.

\bibitem[{Yue et~al.(2024{\natexlab{a}})Yue, Ni, Zhang, Zheng, Liu, Zhang, Stevens, Jiang, Ren, Sun, Wei, Yu, Yuan, Sun, Yin, Zheng, Yang, Liu, Huang, Sun, Su, and Chen}]{yue2023mmmu}
Xiang Yue, Yuansheng Ni, Kai Zhang, Tianyu Zheng, Ruoqi Liu, Ge~Zhang, Samuel Stevens, Dongfu Jiang, Weiming Ren, Yuxuan Sun, Cong Wei, Botao Yu, Ruibin Yuan, Renliang Sun, Ming Yin, Boyuan Zheng, Zhenzhu Yang, Yibo Liu, Wenhao Huang, and 3 others. 2024{\natexlab{a}}.
\newblock Mmmu: A massive multi-discipline multimodal understanding and reasoning benchmark for expert agi.
\newblock In \emph{Proceedings of CVPR}.

\bibitem[{Yue et~al.(2024{\natexlab{b}})Yue, Zheng, Ni, Wang, Zhang, Tong, Sun, Yu, Zhang, Sun, Su, Chen, and Neubig}]{yue2024mmmu}
Xiang Yue, Tianyu Zheng, Yuansheng Ni, Yubo Wang, Kai Zhang, Shengbang Tong, Yuxuan Sun, Botao Yu, Ge~Zhang, Huan Sun, Yu~Su, Wenhu Chen, and Graham Neubig. 2024{\natexlab{b}}.
\newblock Mmmu-pro: A more robust multi-discipline multimodal understanding benchmark.
\newblock \emph{arXiv preprint arXiv:2409.02813}.

\bibitem[{Zhang et~al.(2024{\natexlab{a}})Zhang, Feng, Bai, Du, Hou, Deng, Han, Li, Wang, Liu, Qu, Zhang, Zhao, Liang, Liu, Fang, Yang, Huang, Lin, Zhang, and Ni}]{zhang2024mllmsunderstanddeepimplication}
Chenhao Zhang, Xi~Feng, Yuelin Bai, Xinrun Du, Jinchang Hou, Kaixin Deng, Guangzeng Han, Qinrui Li, Bingli Wang, Jiaheng Liu, Xingwei Qu, Yifei Zhang, Qixuan Zhao, Yiming Liang, Ziqiang Liu, Feiteng Fang, Min Yang, Wenhao Huang, Chenghua Lin, and 2 others. 2024{\natexlab{a}}.
\newblock \href {https://arxiv.org/abs/2410.13854} {Can mllms understand the deep implication behind chinese images?}
\newblock \emph{Preprint}, arXiv:2410.13854.

\bibitem[{Zhang et~al.(2024{\natexlab{b}})Zhang, Du, Chen, Liang, Luo, Zheng, Zhu, Cheng, Xu, Guo et~al.}]{zhang2024cmmmu}
Ge~Zhang, Xinrun Du, Bei Chen, Yiming Liang, Tongxu Luo, Tianyu Zheng, Kang Zhu, Yuyang Cheng, Chunpu Xu, Shuyue Guo, and 1 others. 2024{\natexlab{b}}.
\newblock Cmmmu: A chinese massive multi-discipline multimodal understanding benchmark.
\newblock \emph{arXiv preprint arXiv:2401.11944}.

\bibitem[{Zhang et~al.(2024{\natexlab{c}})Zhang, Li, Zhang, Pu, Cahyono, Hu, Liu, Zhang, Yang, Li et~al.}]{zhang2024lmms}
Kaichen Zhang, Bo~Li, Peiyuan Zhang, Fanyi Pu, Joshua~Adrian Cahyono, Kairui Hu, Shuai Liu, Yuanhan Zhang, Jingkang Yang, Chunyuan Li, and 1 others. 2024{\natexlab{c}}.
\newblock Lmms-eval: Reality check on the evaluation of large multimodal models.
\newblock \emph{arXiv preprint arXiv:2407.12772}.

\bibitem[{Zhang et~al.(2024{\natexlab{d}})Zhang, Jiang, Zhang, Lin, Guo, Qiu, Zhou, Lu, Chang, Qiao et~al.}]{zhang2024mathverse}
Renrui Zhang, Dongzhi Jiang, Yichi Zhang, Haokun Lin, Ziyu Guo, Pengshuo Qiu, Aojun Zhou, Pan Lu, Kai-Wei Chang, Yu~Qiao, and 1 others. 2024{\natexlab{d}}.
\newblock Mathverse: Does your multi-modal llm truly see the diagrams in visual math problems?
\newblock In \emph{European Conference on Computer Vision}, pages 169--186. Springer.

\bibitem[{Zheng et~al.(2024)Zheng, Teng, Yang, Wang, Chen, Gu, Dong, Ding, and Tang}]{zheng2024cogview3}
Wendi Zheng, Jiayan Teng, Zhuoyi Yang, Weihan Wang, Jidong Chen, Xiaotao Gu, Yuxiao Dong, Ming Ding, and Jie Tang. 2024.
\newblock Cogview3: Finer and faster text-to-image generation via relay diffusion.
\newblock In \emph{European Conference on Computer Vision}, pages 1--22. Springer.

\end{thebibliography}

\appendix

\clearpage
\setcounter{page}{1}

\section{Commercial API Support}
\label{sec:commercial_api}
We support mainstream APIs for multimodal tasks. For Vision-Language Models (VLMs), we provide integration with OpenAI, Gemini, Claude, Hunyan, and Qwen. For Text-to-Image (T2I) models, we support DALL-E, Flux, and Kolors. Additionally, we offer OpenAI-style REST API compatibility for both types of tasks, which we highly recommend using for seamless integration and ease of deployment.


\section{Add A New Evaluation Task}
\label{sec:new_task}

This section describes the procedure for adding new evaluation tasks to the benchmark system. The process consists of three main steps:

\subsection{Create Task Configuration}
New evaluation tasks require creating appropriate configuration files in the \texttt{tasks} directory. For simple tasks (e.g., Visual Question Answering), developers can utilize the existing \texttt{VqaBaseDataset} class.

The basic configuration template includes:

\begin{itemize}
    \item \texttt{dataset\_path}: Path to the original dataset
    \item \texttt{split}: Dataset partition (e.g., "image")
    \item \texttt{processed\_dataset\_path}: Storage path for processed datasets (e.g., "CustomBench")
    \item \texttt{processor}: Data processing script (e.g., "process.py")
\end{itemize}

Developers can configure tasks in two ways:

\begin{enumerate}
    \item \textbf{Default Prompt Configuration}: Uses the system's default prompt template ("Answer the question using a single word or phrase.")
    
    \item \textbf{Custom Prompt Configuration}: Allows customization of the prompt template for specific task requirements
\end{enumerate}

\subsection{Implement Data Processing}
Each new task requires a dedicated processing script (specified in the \texttt{processor} field) to transform raw data into the system's standardized format. The script should handle:

\begin{itemize}
    \item Data loading from source files
    \item Format conversion
    \item Quality control checks
    \item Output generation in the expected structure
\end{itemize}

\subsection{Register the Task}
After configuration and processing implementation, the task must be registered in the system's task registry. This involves:

\begin{itemize}
    \item Adding the task to the appropriate configuration files
    \item Updating any necessary dependencies
    \item Verifying integration through test cases
\end{itemize}

The modular design allows for seamless addition of new evaluation tasks while maintaining consistency across the benchmark system.

\section{Benchmarks for VLM evaluation}
Table \ref{tab:vlm_dataset} summarizes the benchmarks utilized by the FlagEval leaderboard for evaluating vision-language models (VLMs). Each benchmark assesses one or more specific model capabilities, such as visual perception, general knowledge, or mathematical reasoning.

\begin{table*}[htb]
\centering
\small
\caption{Evaluation Datasets for Vision-Language Models}
\begin{tabular}{c|cc}
\hline
\textbf{Benchmark} & \textbf{Language} & \textbf{Capability} \\
\hline
Charxiv(Val)\cite{wang2024charxiv} & English & Chart Comprehension \\
CII-Bench(Test)\cite{zhang2024mllmsunderstanddeepimplication} & Chinese & General Knowledge \\
CMMMU(Val)\cite{zhang2024cmmmu} & Chinese & General Knowledge \\
MMMU(Val)\cite{yue2023mmmu} & English & General Knowledge \\
MMMU-Pro(Standard, Vision)\cite{yue2024mmmu} & English & General Knowledge, Visual Perception \\
MathVision(Test)\cite{wang2024mathvision} & English & Mathematical Ability \\
MathVerse(testmini)\cite{zhang2024mathverse} & English & Mathematical Ability \\
MMVET-v2\cite{yu2024mm} & English & General Knowledge, Visual Perception \\
Blink(Val)\cite{fu2024blink} & English & Visual Perception \\
Self-constructed VQA Dataset & English, Chinese & General Knowledge, Visual Perception \\
Self-constructed Text Dataset & English,Chinese & Text Recognition and Understanding \\
\hline
\end{tabular}
\label{tab:vlm_dataset}
\end{table*}

\section{Human Evaluation Process and Scoring Guidelines}
\label{app:human_evaluation}
In this evaluation, images generated by different models from the same textual prompt were simultaneously displayed to annotators in random order and position. Three trained annotators independently rated each image according to specific evaluation dimensions. Annotators sequentially completed scoring for each evaluation dimension before proceeding to the next. After completing scoring for all three dimensions, annotators repeated this process for two additional rounds. The repeated evaluation rounds were designed to measure and ensure the stability and consistency of annotator scoring criteria.

The evaluation dimensions included Text-Image Consistency, image realism, aesthetic quality, and image safety. Text-Image Consistency, realism, and aesthetic quality were scored on a 5-point scale, whereas safety was scored as a binary (0 or 1). Definitions for each evaluation dimension are provided below:

\begin{itemize}
    \item \textbf{Text-Image Consistency}: Assesses the extent to which the generated image accurately reflects the content described by the text.
    \item \textbf{Realism}: Higher realism scores indicate that the image faithfully represents objects' expected shapes and characteristics rather than arbitrary generation. For realistic scenes, the evaluation considers conformity with the real world; for surreal scenes, it considers consistency with common knowledge from animations, sci-fi films, or other surreal contexts.
    \item \textbf{Aesthetic Quality}: Evaluates the aesthetic appeal of the image, including clarity, composition, and alignment with aesthetic standards.
    \item \textbf{Safety}: Ensures the generated images are free from violence, pornography, drug-related content, and anti-social themes.
\end{itemize}

Detailed scoring criteria for each evaluation dimension are summarized in Table \ref{tab:t2i_subject}.

\begin{table*}[htb]
    \centering
    \caption{Scoring Guidelines for Human Evaluation of Text-to-Image Models}
    \renewcommand{\arraystretch}{1.5}
    \resizebox{\linewidth}{!}{
    \begin{tabular}{p{1.5cm}|p{4cm}|p{4cm}|p{4cm}|p{4cm}}
        \toprule
        \textbf{Score} & \textbf{Text-Image Consistency} & \textbf{Realism} & \textbf{Aesthetic Quality} & \textbf{Safety} \\
        \midrule
        0 & — & — & — & Image contains unsafe content \\
        1 & Image does not reflect textual description & Significant errors in shapes and characteristics & Subjects difficult to identify; chaotic composition & Image is safe \\
        2 & Image includes a few elements from the textual description & Some noticeable errors in shapes and characteristics & Poor aesthetic quality & — \\
        3 & Image contains most elements of the textual description & Some obvious errors but overall coherent & Average aesthetic quality; ordinary composition and color matching & — \\
        4 & Image reflects textual description well & Minor, less obvious errors in shapes and characteristics & Good aesthetics with slight shortcomings in composition or color matching & — \\  
        5 & Image perfectly aligns with textual description & No errors; image is coherent and realistic & Excellent aesthetic quality with outstanding composition and color matching & — \\  
        \bottomrule
    \end{tabular}
    }
    \label{tab:t2i_subject}
\end{table*}

\end{document}